\documentclass[twocolumn]{article} % For LaTeX2e

\usepackage{graphicx} % more modern
\usepackage{subfigure}
\usepackage{amssymb}
\usepackage{amsmath}
\usepackage{tweaklist}
\usepackage{algorithm}
\usepackage{algorithmic}

\usepackage{hyperref}

\newlength\titlebox 
\def\bi{\begin{itemize}}
\def\ei{\end{itemize}}
\def\L{{\mathcal{L}}}
\def\N{{\mathcal{N}}}
\def\beqas{\begin{eqnarray*}}
\def\eeqas{\end{eqnarray*}}
\def\beqa{\begin{eqnarray}}
\def\eeqa{\end{eqnarray}}
\def\beq{\begin{equation}}
\def\eeq{\end{equation}}
\def\bit{\begin{itemize}}
\def\eit{\end{itemize}}
\def\ben{\begin{enumerate}}
\def\een{\end{enumerate}}
\def\BA{\begin{array}}
\def\EA{\end{array}}
\def\diag{\mathop{\rm diag}}

\def\chol{\mathop{\rm chol}}

\def\Diag{\mathop{\rm Diag}}
\def\x{\mathbf{x}}
\def\s{\mathbf{s}}

\newcommand{\Cov}{{\mathop { \rm cov{}}}}

\newcommand{\tr}{\mathop{ \rm tr}}

\newcommand{\idm}{I}
\newcommand{\rb}{\mathbb{R}}
\newcommand{\BlackBox}{\rule{1.5ex}{1.5ex}}  % end of proof

\newenvironment{proof}{\par\noindent{\bf Proof\ }}{\hfill\BlackBox\\[2mm]}

\newtheorem{proposition}{Proposition}

\title{Local Component Analysis}
\author{
Nicolas Le Roux\\
\texttt{nicolas@le-roux.name}
\and
Francis Bach\\
\texttt{francis.bach@ens.fr}
\and
\\
INRIA - SIERRA Project - Team\\
Laboratoire d'Informatique de l'\'Ecole Normale Sup\'erieure\\
Paris, France
}

\date{}

\begin{document}

\maketitle
\begin{abstract}
Kernel density estimation, a.k.a.~Parzen windows, is a popular density
estimation method,  which can be used for outlier detection or clustering.
With multivariate data, its performance is heavily reliant on the metric
used within the kernel. Most earlier work has focused on learning only the
bandwidth of the kernel (i.e., a scalar multiplicative factor). In this
paper, we propose to learn a full Euclidean metric through an
expectation-minimisation (EM) procedure, which can be seen as an
unsupervised counterpart to neighbourhood component analysis (NCA). In order to
avoid overfitting with a fully nonparametric density estimator in high
dimensions, we also consider a semi-parametric Gaussian-Parzen density
model, where some of the variables are modelled through a jointly Gaussian
density, while others are modelled through Parzen windows. For these two
models, EM leads to simple closed-form updates based on matrix inversions and
eigenvalue decompositions. We show empirically that our method leads to
density estimators with higher test-likelihoods than natural competing
methods, and that the metrics may be used within most unsupervised learning
techniques that rely on local distances, such as spectral clustering or manifold
learning methods. Finally, we present a stochastic approximation scheme which allows for the use of this method in a large-scale setting.
\end{abstract}

\section{Introduction}
Most unsupervised learning methods rely on a metric on the space of
observations. The quality of the metric directly impacts the
performance of such techniques and a significant amount of work has
been dedicated to learning this metric from data when some supervised
information is available~\cite{Xing02,Goldberger04,Bach2003_learningSC}. However, in a fully
unsupervised scenario, most practitioners use the Mahalanobis
distance obtained from principal component analysis (PCA). This is an
unsatisfactory solution as PCA is essentially a global linear
dimension reduction method, while most unsupervised learning
techniques, such as spectral clustering or manifold learning, are local.

In this paper, we cast the unsupervised metric learning as a density
estimation problem with a Parzen windows estimator based on a
Euclidean metric. Using the maximum likelihood framework, we derive in
Section~\ref{sec:em_procedure} an expectation-minimisation (EM) procedure that maximizes the
\emph{leave-one-out log-likelihood}, which may be considered as an
unsupervised counterpart to neighbourhood component
analysis (NCA)~\cite{Goldberger04}. As opposed to PCA, which performs a
whitening of the data based on global information, our new algorithm
globally performs a whitening of
the data using only local information, hence the denomination local component analysis (LCA).

Like all non-parametric density estimators, Parzen windows density
estimation is known to overfit in high dimensions~\cite{Silverman86}, and thus LCA should
also overfit. In order to keep the
modelling flexibility of our density estimator while avoiding
overfitting, we propose a semi-parametric Parzen-Gaussian model;
following~\cite{Blanchard06}, we linearly transform then split our variables
in two parts, one which is modelled through a Parzen windows estimator
(where we assume the interesting part of the data lies), and one which is
modelled as a multivariate Gaussian (where we assume the noise lies). Again,
in Section~\ref{sec:mult_gauss_components}, an EM procedure for estimating
the linear transform may be naturally derived and leads to simple
closed-form updates based on matrix inversions and eigenvalue
decompositions. This procedure contains no hyperparameters, all the
parameters being learnt from data.

Since the EM formulation of LCA scales quadratically in the number of
datapoints, making it impractical for large datasets, we introduce in
Section~\ref{sec:subsampling} both a stochastic approximation and a
subsampling technique allowing us to achieve a linear cost and thus to scale
LCA to much larger datasets.

Finally, in Section~\ref{sec:experiments}, we show empirically that our
method leads to
density estimators with higher test-likelihoods than natural competing
methods, and that the metrics may be used within unsupervised learning
techniques that rely on such metrics, like spectral clustering.

\section{Previous work}
Many authors aimed at learning a Mahalanobis distance suited for local learning. While some techniques required the presence of labelled data~\cite{Goldberger04,Xing02,Bach2003_learningSC}, others proposed ways to learn the metric in a purely unsupervised way, e.g.,~\cite{zelnik2004self} who used the distance to the $k$-th nearest neighbour as the local scaling around each datapoint. Most of the other attempts at unsupervised metric learning were developed in the context of kernel density estimation, a.k.a. Parzen windows. The Parzen windows estimator~\cite{Parzen62} is a nonparametric density estimation model
which, given $n$ datapoints $\{\x_1,
\ldots, \x_n\}$ in $\mathbb{R}^d$, defines a mixture model of the form $p(\x) = \frac{1}{n} \sum_{j=1}^n K(\x, \x_j, \theta)$
where $K$ is a kernel with compact support and parameters $\theta$. We
relax the compact support assumption and choose $K$ to be the
normal kernel, that is
\begin{align*}
p(\x) &= \frac{1}{n} \sum_{j=1}^n \N(\x, \x_j, \Sigma)\\
	&\propto \frac{1}{n\sqrt{|\Sigma|}}\sum_{j=1}^n \exp\left[-\frac12(\x - \x_j)^\top\Sigma^{-1}(\x - \x_j)\right] \; ,
\end{align*}
where $\Sigma$ is the covariance matrix of each Gaussian. As the performance of the Parzen windows estimator is more reliant on the covariance matrix than on the kernel, there has been a large body of work, originating from the statistics literature, attempting to learn this matrix. However, almost all attempts are focused on the asymptotic optimality of the estimators obtained with little consideration for the practicality in high dimensions. Thus, the vast majority of the work is limited to isotropic matrices, reducing the problem to finding a single scalar $h$~\cite{Rosenblatt56,Duin76,Rudemo82,Chow83,Bowman84,Park90,Sheather91}, the \textit{bandwidth}, and the few extensions to the non-isotropic cases are numerically expensive~\cite{Duong2005,Jones09}.

An exception is the approach proposed in~\cite{Vincent02manifoldparzen}, which is very similar to our method, as the authors learn the covariance matrix of the Parzen windows estimator using local neighbourhoods. However, their algorithm does not minimize a well-defined cost function, making it unsuitable for kernels other than the Gaussian one, and the locality used to compute the covariance matrix depends on parameters which must be hand-tuned or cross-validated. Also, the modelling of all the dimensions using the Parzen windows estimator makes the algorithm unsuitable when the data lie on a high-dimensional manifold. In an extension to~\cite{Vincent02manifoldparzen}, ~\cite{Bengio06non-localmanifold} uses a neural network to compute the leading eigenvectors of the local covariance matrix at every point in the space, then uses these matrices to do density estimation and classification. Despite the algorithm's impressive performance, it does not correspond to a linear reparametrisation of the space and thus cannot be used as a preprocessing step.

\section{Local Component Analysis}
\label{sec:em_procedure}

Seeing the density as a mixture of Gaussians, one can easily optimize the
covariances using the EM algorithm~\cite{Dempster77}. However, maximizing
the standard log-likelihood of the data would trivially lead to the degenerate solution where $\Sigma$ goes
to 0 to yield a sum of Dirac distributions. One solution to that problem is
to penalize some norm of the precision matrix to prevent it from going to
infinity. Another, more compelling, way is to optimize the leave-one-out
log-likelihood, where the probability of each datapoint $\x_i$ is computed
under the distribution obtained when $\x_i$ has been
removed from the training set. This technique is not new and has already
been explored both in the supervised~\cite{Goldberger04,Globerson06} and in
the unsupervised setting~\cite{Duin76}. However, in the latter case, the
cross-validation was then done by hand, which explains why only one
bandwidth parameter
could be optimized\footnote{Most of the literature on estimating the
covariance matrix discards the log-likelihood cost because of its
sensitivity to outliers and prefers AMISE (see, e.g.,~\cite{Duong2005}). However, in all our experiments,
the number of datapoints was large enough so that LCA did not suffer from
the presence of outliers.}. We will thus use the following criterion:
\begin{align}
	\L(\Sigma) &= -\sum_{i=1}^n \log\Big[\frac{1}{n-1}\sum_{j\neq i}
\N\left(x_i, x_j, \Sigma\right)\Big]\label{eq:sigma_cost}\\
&\leq \mbox{cst} - \sum_{i=1}^n \sum_{j\neq i} \lambda_{ij}\log\N\left(x_i,
x_j, \Sigma\right)\nonumber\\
&\hspace*{2.2cm} + \sum_{i=1}^n
\sum_{j\neq i} \lambda_{ij}\log \lambda_{ij} \; ,\label{eq:after_jensen}
\end{align}
with the constraints $\forall i \; , \; \sum_{j\neq i} \lambda_{ij} = 1$. This variational bound is obtained using Jensen's inequality.

The EM algorithm optimizes the right-hand side of
Eq.~(\ref{eq:after_jensen}) by alternating between the optimisations of
$\lambda$ and $\Sigma$ in turn. The algorithm is guaranteed to converge,
and does so to a stationary point of the true function over $\Sigma$ defined in Eq.~(\ref{eq:sigma_cost}). At each step, the optimal solutions are:
\begin{eqnarray}
	\lambda_{ij}^* &=& \frac{\N(x_i, x_j, \Sigma)}{\sum_{k\neq i}
\N(x_i, x_k, \Sigma)} \text{ if } j \neq i\label{eq:lambda_opt_diff}\\
	 \lambda_{ii}^* &=& 0\label{eq:lambda_opt_equal}\\
	\Sigma^* &=& \frac{\sum_{ij} \lambda_{ij} (\x_i - \x_j)(\x_i -
\x_j)^T}{n} \; .
\label{eq:optimal_sigma}
\end{eqnarray}

The ``responsibilities'' $\lambda^*_{ij}$ define the relative proximity of $\x_j$ to $\x_i$
(compared to the proximity of all the $\x_k$'s to $\x_i$) and $\Sigma^*$ is the average of all the local covariance matrices.

\begin{figure*}[t]
\begin{center}
\begin{tabular}{ccc}
\includegraphics[height=1cm]{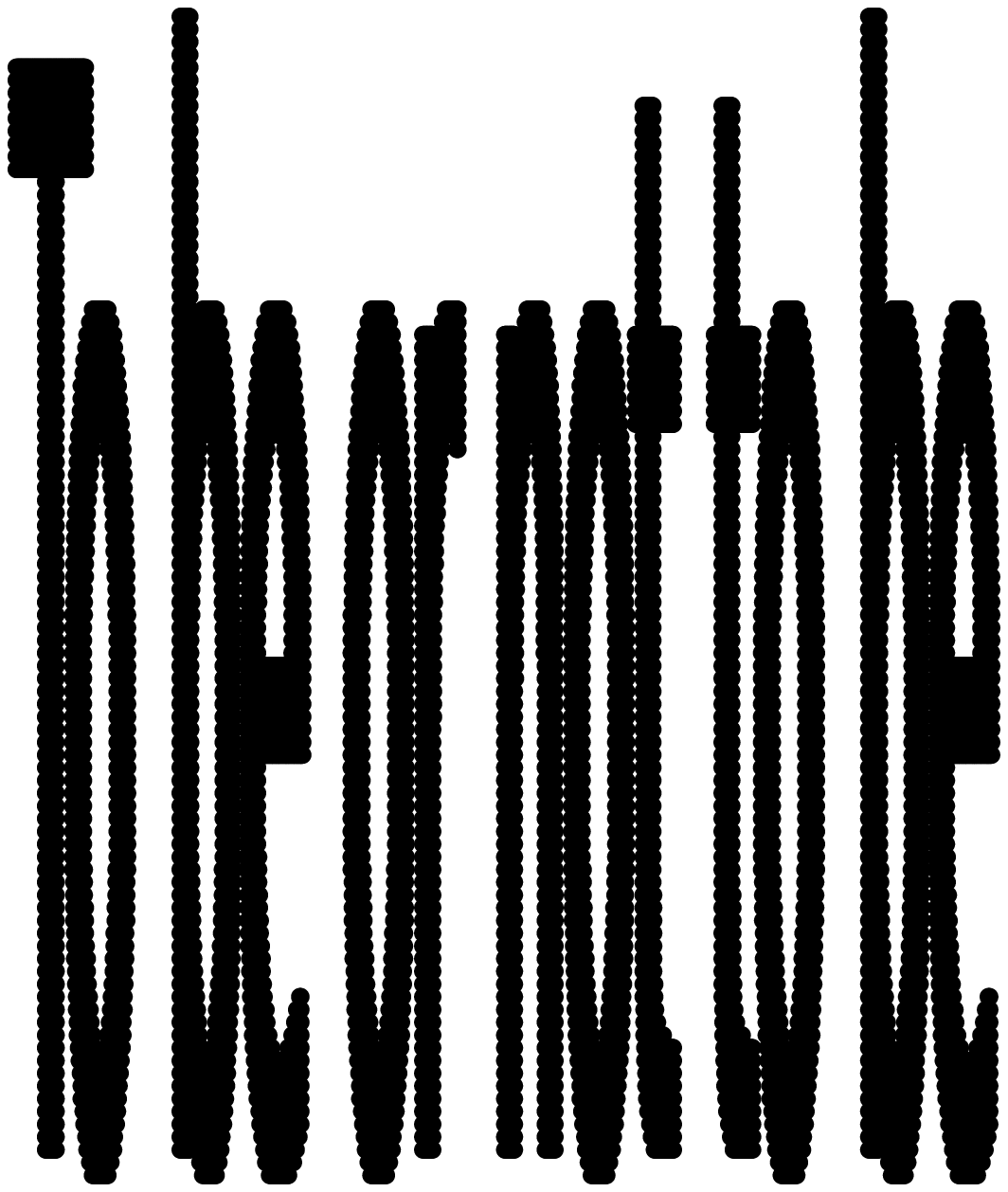}&\hspace*{1cm}&
	\includegraphics[height=1cm]{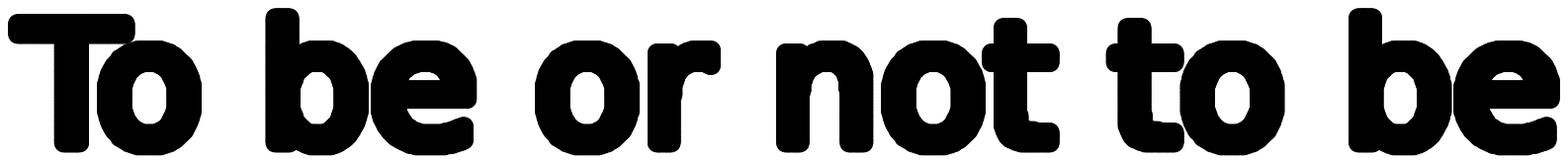}
\end{tabular}
	\caption{Results obtained when transforming ``To be or not to be'' using PCA (left)
and LCA (right).
\textbf{Left}: To make the data globally
isotropic, PCA awkwardly compresses the letters horizontally.
\textbf{Right}: Since LCA is insensitive to the spacing between letters and to the length of the
text, there is no horizontal compression.
	\label{fig:nips_pca_lca}}
\end{center}
\end{figure*}

This algorithm, which we coin LCA, for local component analysis, transforms
the data to make it locally isotropic, as opposed to PCA which makes it globally isotropic. Fig.~(\ref{fig:nips_pca_lca}) shows a comparison
of PCA and LCA on the word sequence ``To be or not to be''. Whereas PCA is highly
sensitive to the length of the text, LCA is only affected by the local
shapes, thus providing a much less distorted result.\footnote{Since both methods are insensitive to any linear reparametrisation of the data, we do not include the original data in the figure.}

First, one may note at this point that Manifold Parzen Windows~\cite{Vincent02manifoldparzen} is equivalent to LCA with only one step of EM. This makes Manifold Parzen Windows more sensitive to the choice of the original covariance matrix whose parameters must be carefully chosen. As we shall see later in the experiments, running EM to convergence is important to get good accuracy when using spectral clustering on the transformed data. Second, it is also worth noting that, similarly to Manifold Parzen Windows, LCA can straightforwardly be extended to the cases where each datapoint uses its own local covariance matrix (possibly with a smoothing term), or where the covariance $\Sigma^*$ is the sum of a low-rank matrix and some scalar multiplied by the identity matrix.

Not only may LCA be used to learn a linear transformation of the space, but
it also defines a density model. However, there are two potential negative
aspects associated with this method. First, in high dimensions, Parzen
windows is prone to overfitting and must be regularized~\cite{Silverman86}.
Second, if there are some directions containing a small Gaussian noise, the
local isotropy will blow them up, swamping the data with clutter. This is
common to all the techniques which renormalise the data by the inverse of
some variance. A solution to both of these issues is to consider a product
of two densities: one is a low-dimensional Parzen windows estimator, which
will model the interesting signal, and the other is a Gaussian, which will
model the noise.

\section{LCA with a multiplicative Gaussian component}
\label{sec:mult_gauss_components}
We now assume that there are irrelevant dimensions
in our data which can be modelled by a Gaussian. In other words,  we
consider an invertible linear transformation $(B_G, B_L)^\top \x$ of the data, modelling
$B_G^\top \x $ as a multivariate Gaussian and $B_L^\top \x$ through
kernel density estimation,  the two parts being independent, leading
to $p(\x) \propto p(B_G^{\top}\x, B_L^{\top}\x) = p_G(B_G^{\top}\x)p_L(B_L^{\top}\x)$, where $p_G$ is a Gaussian and $p_L$ is the Parzen windows estimator, i.e.,
\begin{align*}
p(\x_i) &\propto
\frac{\left|B_GB_G^\top+B_LB_L^\top\right|^{\frac{1}{2}}}{n-1}\\
&\hspace*{.5cm} \times \exp\left[-\frac{1}{2}(\x_i - \mu)^\top B_GB_G^\top
(\x_i-\mu)\right]\\
&\hspace*{.5cm} \times \left(\sum_{j\neq i} \exp\left[-\frac{1}{2}(\x_i - \x_j)^\top B_LB_L^\top (\x_i-\x_j)\right]\right) \; ,
\end{align*}
with $(B_G, B_L)$ a full-rank square matrix. Using EM, we can upper-bound the negative log-likelihood:
\begin{align}
-2 \sum_i \log p(\x_i) &\leq \tr(B_G^\top C_GB_G) + \tr(B_L^\top C_LB_L)\nonumber\\
	&\hspace*{1cm} -\log |B_GB_G^\top + B_L B_L^\top| \; ,
\label{eq:mult_gauss_up_bound}
\end{align}
with
\begin{align*}
C_G &= \frac{1}{n} \sum_i (\x_i - \mu)(\x_i - \mu)^\top \;,\\
C_L &= \frac{1}{n} \sum_{ij}\lambda_{ij} (\x_i - \x_j)(\x_i - \x_j)^\top \; .
\end{align*}
The matrices $B_G$ and $B_L$ minimizing the right-hand side of
Eq.~(\ref{eq:mult_gauss_up_bound}) may be found using the following
proposition (see proof in the appendix):
\begin{proposition}
\label{prop:mult_gauss}
Let $B_G \in \rb^{d \times d_1}$ and $B_L \in \rb^{d \times d_2}$, with
$d=d_1+d_2$ and $B=(B_G,B_L) \in \rb^{d \times d}$ invertible. Consider two
symmetric positive matrices $M_1$ and $M_2$ in $\rb^{d \times d}$.
The problem
\beq
\min_{B_G, B_L}\ \  \tr B_G^\top M_1 B_G + \tr B_L^\top M_2 B_L
- \log \det ( B_G B_G^\top + B_L B_L^\top)
\eeq
has a finite solution only if $M_1$ and $M_2$ are invertible and, if these conditions are met, reaches
its optimum at
\[
	B_G = M_1^{-1/2}U_+ \; , \quad B_L = M_1^{-1/2}U_- D_-^{-1/2} \; ,
\]
where $U_+$ are the eigenvectors of $M_1^{-1/2}M_2M_1^{-1/2}$ associated
with eigenvalues greater than or equal to 1, $U_-$ are the eigenvectors of
$M_1^{-1/2}M_2M_1^{-1/2}$ associated with eigenvalues smaller than 1 and
$D_-$ is the diagonal matrix containing the eigenvalues of
$M_1^{-1/2}M_2M_1^{-1/2}$ smaller than 1.
\end{proposition}

The resulting procedure is described in Algorithm~\ref{alg:lca_gauss}, where
all dimensions are initially modelled by the Parzen windows estimator, which
empirically yielded the best results.

\begin{algorithm}
\caption{LCA - Gauss\label{alg:lca_gauss}}
\begin{algorithmic}
\REQUIRE $X$ (dataset), iterMax (maximum number of iterations), $\nu$ (regularisation)
\ENSURE $B_G$ (Gaussian part transformation), $B_L$ (Parzen windows transformation)
\STATE $C_G \gets \Cov(X) + \nu I_d\;$ \COMMENT{Initialize $C$ to the global covariance}
\STATE $I_G \gets C_G^{-\frac12}$
\STATE $B_G = 0, B_L = \chol(C_G^{-1})\;$\COMMENT{Assign all dimensions to the Parzen windows estimator}
\FOR{iter = 1:iterMax}
	\STATE $M_{ij} \gets \exp\left[-\frac{(x_i - x_j)^\top B_L^\top B_L (x_i - x_j)}{2}\right]$, $\quad M_{ii} \gets 0$
	\STATE $\lambda_{ij} \gets \frac{M_{ij}}{\sum_k M_{ik}}$
	\STATE $C_L \gets \frac{\sum_{ij}\lambda_{ij}(x_i - x_j)(x_i - x_j)^\top}{n} + \nu I_d$
	\STATE $[V,D] \gets eig(I_GC_LI_G)\;$\COMMENT{Eigendecomposition of $I_GC_LI_G$}
	\STATE $t_1 = \max_z D(z, z) \leq 1\;$\COMMENT{Cut-off between eigenvalues smaller and larger than 1}
	\STATE $t_+ = \{t | t_1 \leq t \leq d\}$ , $t_- = \{t | 1 \leq t < t_1\}$
	\STATE $V_+ \gets V(:, t_+)$, $\quad V_- \gets V(:, t_-)$, $\quad D_- = D(t_-, t_-)$
	\STATE $B_L = I_GV_-D_-^{-1/2}$, $\quad B_G = I_G V_+$
\ENDFOR
%\RETURN $C$
\end{algorithmic}
\end{algorithm}

\paragraph{Relationship with ICA.}
Independent component analysis (ICA) can be seen as a density model
where $\x = A\s$ and $\s$ has independent components (see, e.g.,~\cite{ICA-book}).
In the Parzen windows framework, this corresponds to modelling the
density of $\s$ by a product of univariate kernel density
estimators~\cite{boscolo2004independent}. This however causes two
problems: first, while this assumption is appropriate in settings such as
source separation, it is violated in most settings, and having a
multivariate kernel density estimation is preferable. Second, most
algorithms are dedicated to finding independent components which are
\emph{non-Gaussian}. In the presence of more than one Gaussian dimension, most ICA
frameworks become unidentifiable, while our explicit modelling of such
Gaussian components allows us to tackle this situation (a detailed
analysis of the identifiability of our Parzen/Gaussian model is out of
the scope of this paper).

\paragraph{Relationship with NGCA.}
NGCA~\cite{Blanchard06} makes an assumption similar to ours (they rather assume an additive Gaussian noise on top of a low-dimensional non-Gaussian signal) but uses a projection pursuit algorithm to iteratively
find the directions of non-Gaussianity. Unlike in FastICA, the contrast functions
used to find the interesting directions can be different for each direction.
However, like all projection pursuit algorithms, the identification of
interesting directions gets much harder in higher dimensions, as most of
them will be almost Gaussian. Our use of a non-parametric density estimator
with a log-likelihood cost allows us to globally optimize all directions
simultaneously and does not rely on the model being correct. Finally, LCA estimates all its parameters from data as opposed
to NGCA which requires the number of non-Gaussian directions to be set.

\paragraph{Escaping local optima.}
Though our model allows for the modification of the number of dimensions
modelled by the Gaussian through the analysis of the spectrum of
$C_G^{-1/2}C_LC_G^{-1/2}$, it is sensitive to local optima. It is for
instance rare that a dimension modelled by a Gaussian is switched to the
Parzen windows estimator. Even though the algorithm will more easily
switch from the Parzen windows estimator to the Gaussian model, it will
typically stop too early, that is model many dimensions using the Parzen
windows estimator rather than the better Gaussian. To solve these issues, we propose an alternate algorithm, \emph{LCA-Gauss-Red}, which explores the space of dimensions modelled by a Gaussian more aggressively using a search algorithm, namely:
\begin{enumerate}
\item We run the algorithm LCA - Gauss for a few iterations (40 in our experiments);
\item We then ``transfer'' some columns from $B_L$ (the Parzen windows model) to $B_G$ (the Gaussian model), and rerun LCA - Gauss using these new matrices as initialisations;
\item We iterate step 2 using a dichotomic search of the optimal number of dimensions modelled by the Gaussian, until a local optimum is found;
\item Once we have a locally optimum number of dimensions modelled by the Gaussian model, we run LCA - Gauss to convergence.
\end{enumerate}
\section{Speeding up LCA}
\label{sec:subsampling}
Computing the local covariance matrix of the points using
Eq.~(\ref{eq:lambda_opt_diff}),~(\ref{eq:lambda_opt_equal}) and~(\ref{eq:optimal_sigma}) has a complexity
in $O(dn^2 + d^2n + d^3)$, with $d$ the dimensionality of the data and $n$
the number of training points. Since this is impractical for large datasets,
we can resort to sampling to keep the cost linear in the number of
datapoints. We may further use low-rank or diagonal approximation to achieve
a complexity which grows quadratically with $d$ instead of cubically.

\subsection{Averaging a subset of the local covariance matrices}
Instead of averaging the local covariances over all datapoints, we may only
average them over a subset of datapoints. This estimator is unbiased and, if the local covariance matrices are not too dissimilar, which is the assumption underlying LCA, then its variance should remain small. This is equivalent to using a minibatch procedure: every time we have a new
minibatch of size B, we compute its local covariance $\widehat{C_L}$, which
is then averaged with the previously computed $C_L$ using
\beq
C_L \leftarrow \gamma^{\frac{B}{n}} C_L + (1-\gamma^{\frac{B}{n}}) \widehat{C_L}
\eeq
to yield the updated $C_L$. The exponent $B/n$ is so that $\gamma$, the discount factor,
determines the weight of the old covariance matrix after an entire pass
through the data, which makes it insensitive to the particular choice of
batch size. As opposed to many such algorithms where the choice of $\gamma$
is critical as it helps retaining the information of previous batches, the
locality of the EM estimate makes it less so. However, if the number of
datapoints used to estimate $C_L$ is not much larger than the dimension of
the data, we need to set a higher $\gamma$ to avoid degenerate covariance
matrices. In simulations, we found that using a value of $\gamma = .6$
worked well. Similarly, the size of the minibatch influences only marginally
the final result and we found a value of 100 to be large enough.

\subsection{Computing the local covariance matrices using a subset of the datapoints}
Rather than using only a subset of local covariance matrices, one may also
wonder if using the entire dataset to compute these matrices is necessary.
Also, as the number of datapoints grows, the chances of overfitting
increase. Thus, one may choose to use only a subset of the datapoints to
compute these matrices. This will increase the local covariances, yielding a
biased estimate of the final result, but may also act as a regulariser. In
practice, for very large datasets, one will want the largest neighbourhood
size while keeping the computational cost tractable.

Denoting $n_i$ the number of locations at which we estimate the local
covariance and $n_j$ the number of neighbours used to estimate this
covariance, the cost per update is now $O(d^2 [n_i + n_j] + dn_in_j +
d^3)$. Since only $n_j$ should grow with $n$, this is
linear in the total number of datapoints.

Though they may appear similar, these are not ``landmark'' techniques (see, e.g.,~\cite{DeSilva04}) as
there is still one Gaussian component per datapoint, and the $n_i$ datapoints around which we compute the local covariances are randomly sampled at every iteration.

\section{Experiments}
\label{sec:experiments}
LCA has three main properties: first, it transforms the data to make it
locally isotropic, thus being well-suited for preprocessing the data before
using a clustering algorithm like spectral clustering; second, it
extracts relevant, non-Gaussian components in the data; third, it
provides us with a good density model through the use of the Parzen windows
estimator.

In the experiments, we will assess the performance of the following
algorithms: \emph{LCA}, the original algorithm; \emph{LCA-Gauss}, using a
multiplicative Gaussian component, as described in
Section~\ref{sec:mult_gauss_components}; \emph{LCA-Gauss-Red}, the variant
of \emph{LCA-Gauss} using the more aggressive search to find a better
number of dimensions to be modelled by the Gaussian component. The MATLAB code
for \emph{LCA}, \emph{LCA-Gauss} and \emph{LCA-Gauss-Red} is available at \url{http://nicolas.le-roux.name/code.html}.

\subsection{Improving clustering methods}
\begin{figure*}[ht!]
\begin{center}
	\includegraphics[width=3.5cm]{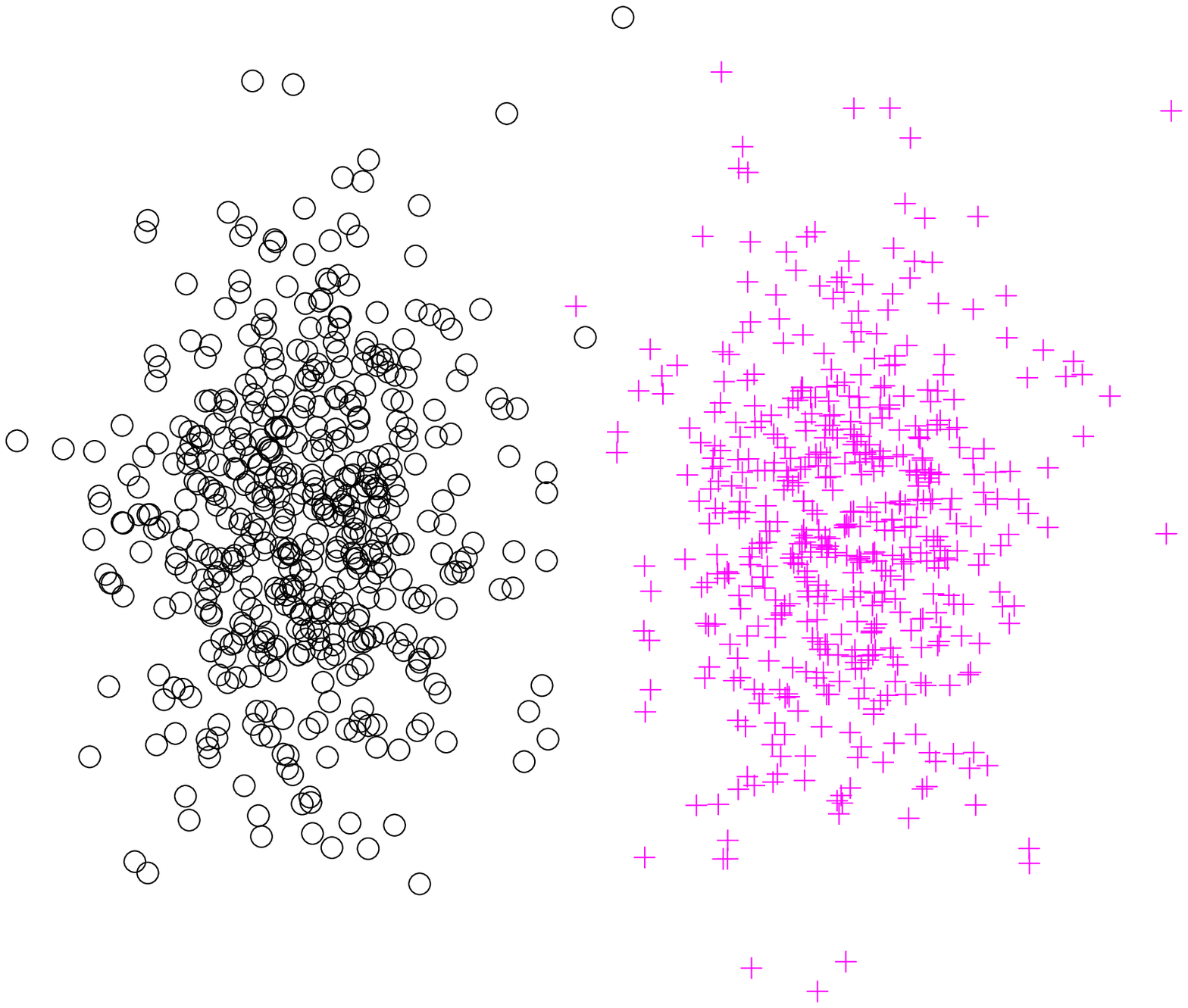}\hspace*{1cm}
	\includegraphics[width=3.5cm]{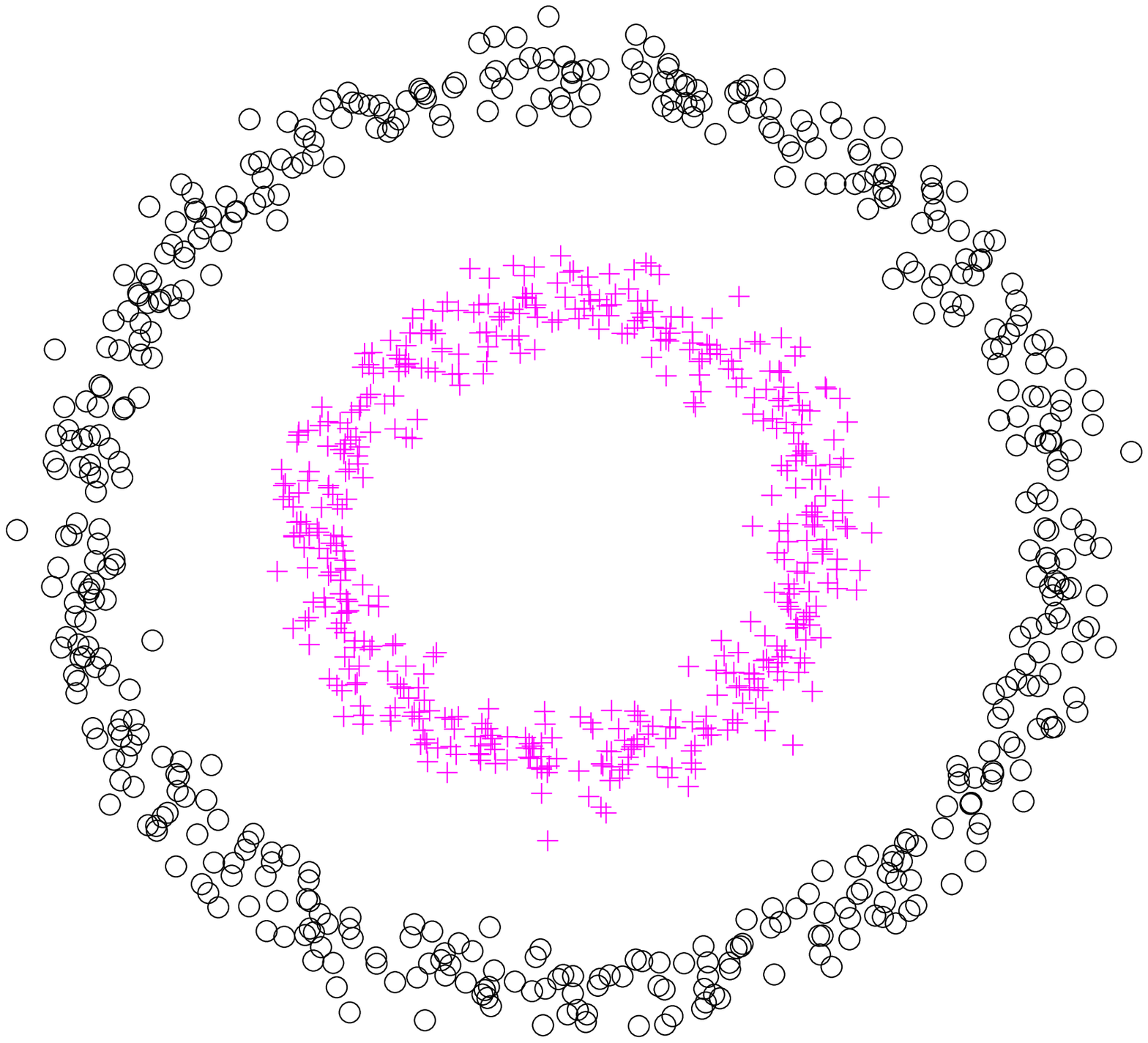}\hspace*{1cm}
	\includegraphics[width=3.5cm]{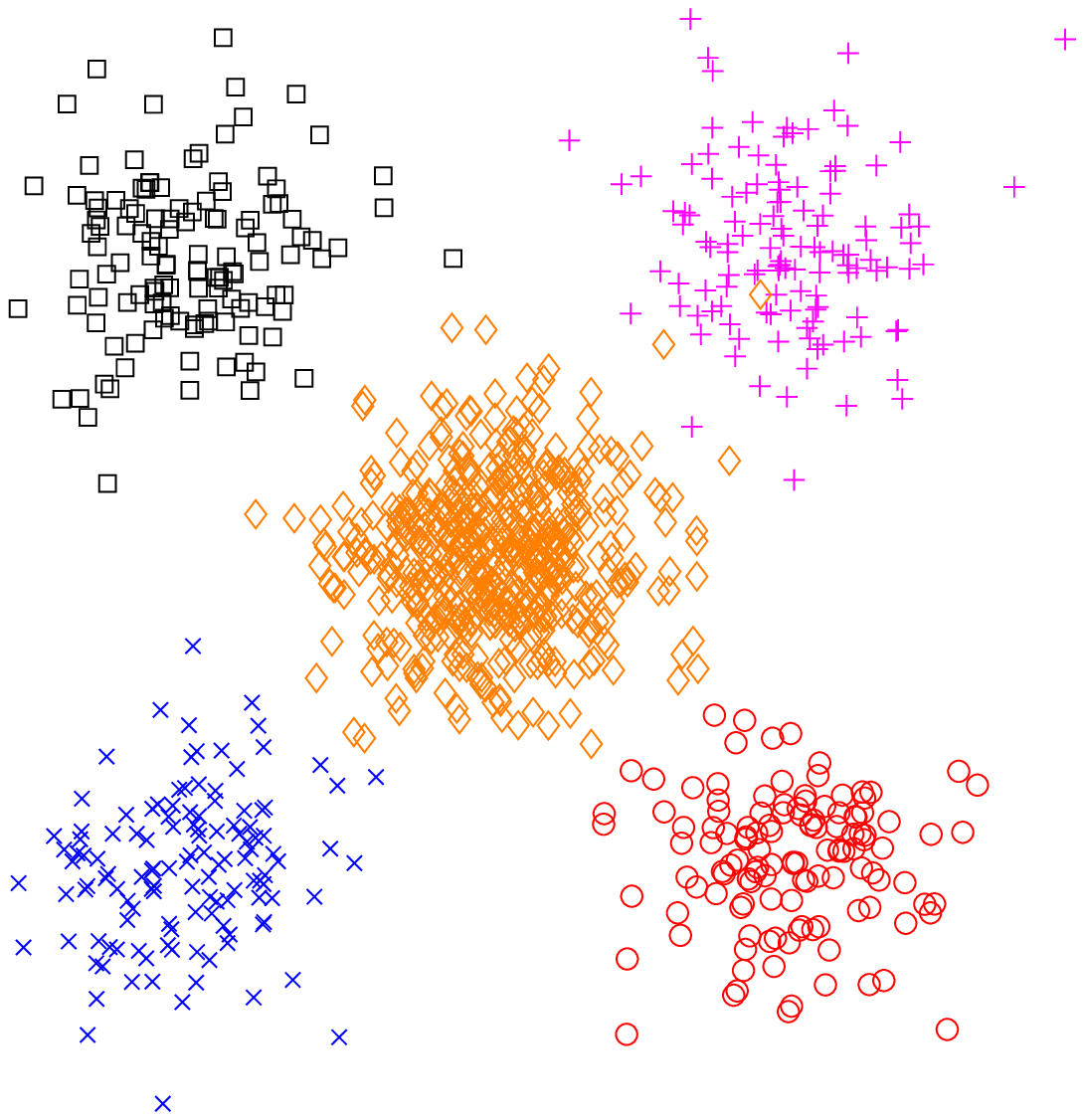}
	\caption{Noise-free data used to assess the robustness of $K$-means
to noise. Left: mixture of two isotropic Gaussians of unit variance and
means $[-3, 0]^\top$ and $[3, 0]^\top$. Centre: two concentric circles
with radii 1 and 2, with added isotropic Gaussian noise of standard
deviation $.1$. Right: mixture of five Gaussians. The centre cluster
contains four times as many datapoints as the other ones.
	\label{fig:two_blob_data}}
\end{center}
\end{figure*}

\begin{figure*}[ht!]
\begin{center}
	\includegraphics[width=4.2cm]{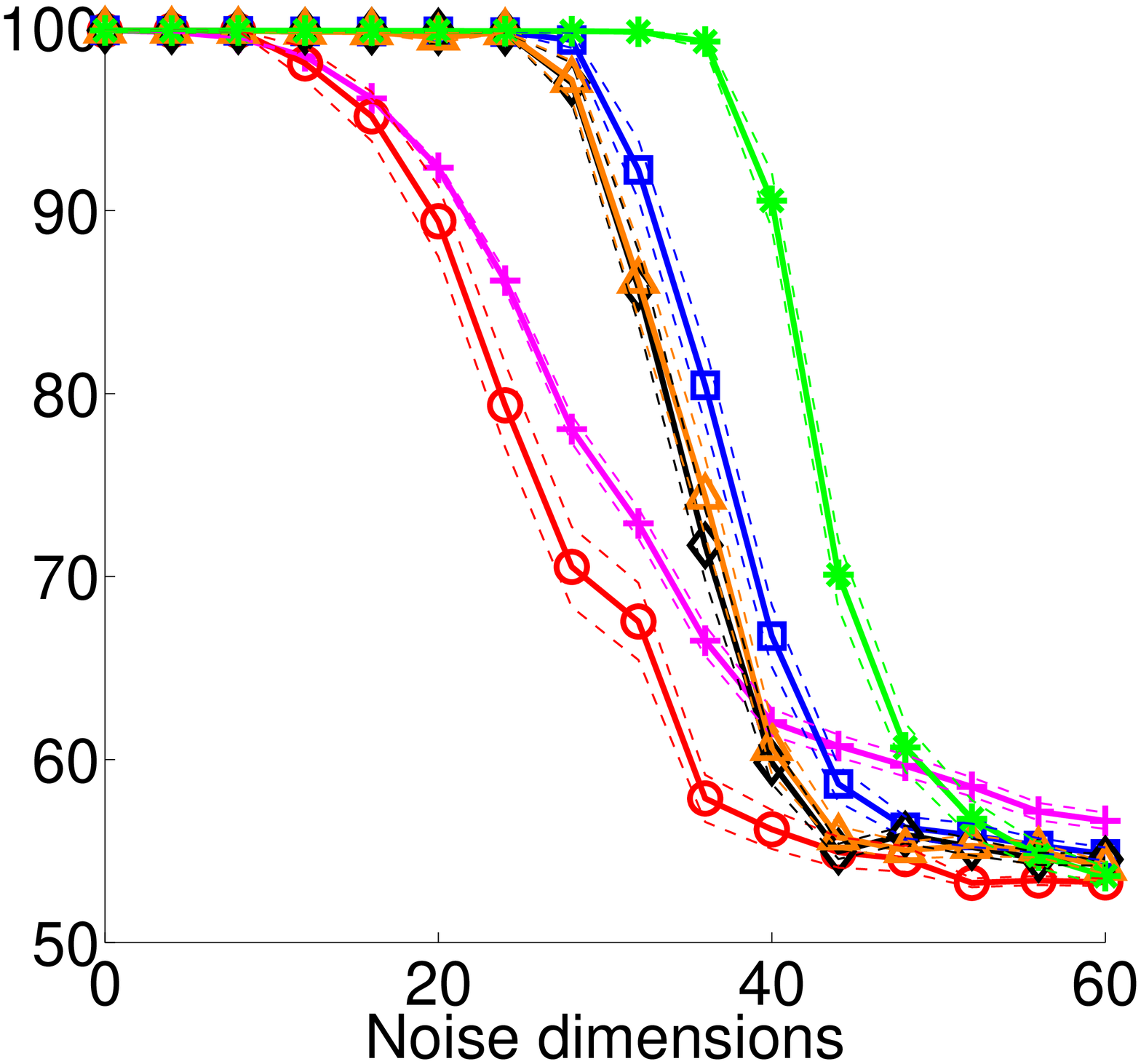}\hspace*{10mm}
	\includegraphics[width=4.2cm]{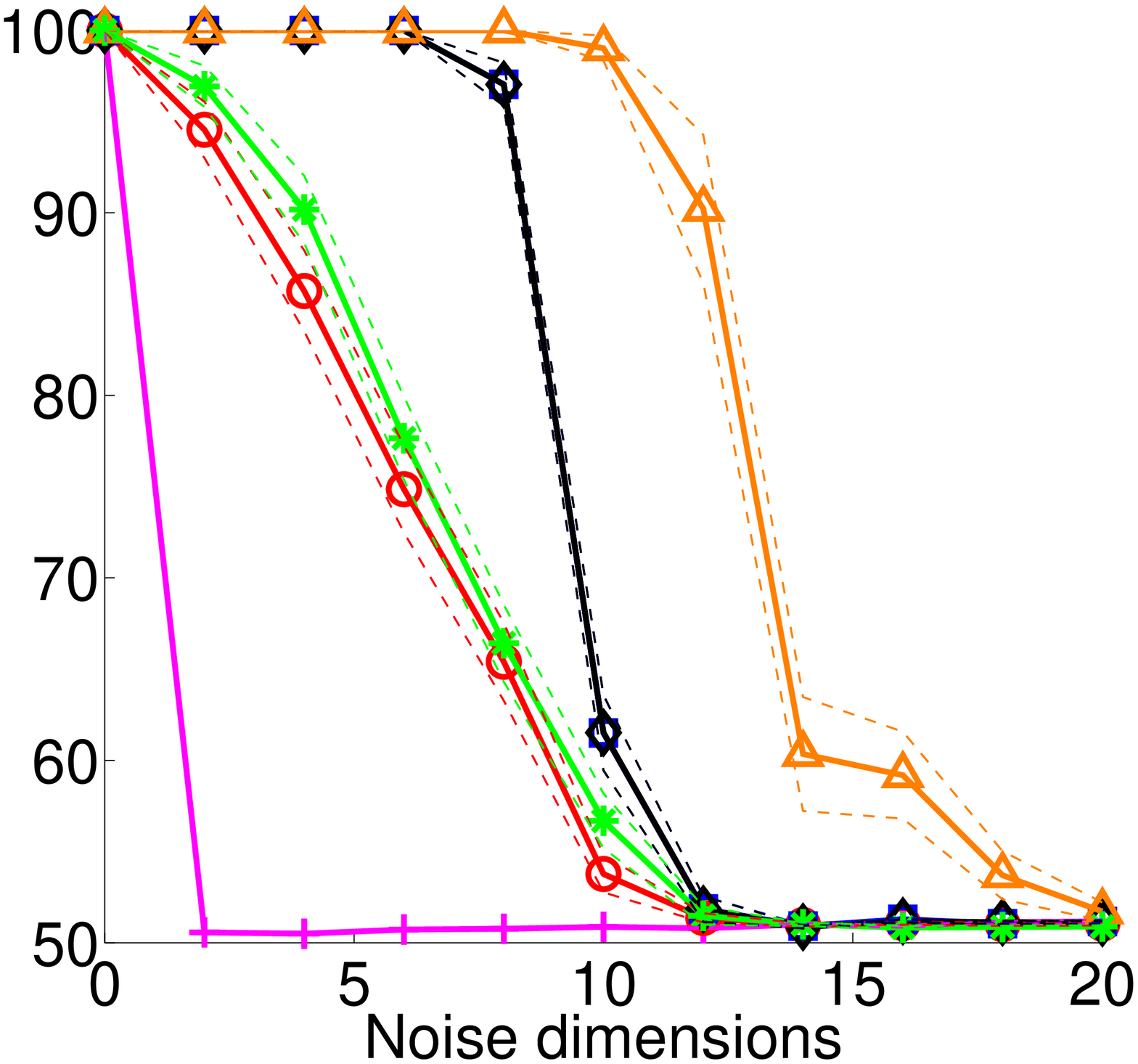}\hspace*{10mm}
	\includegraphics[width=4.2cm]{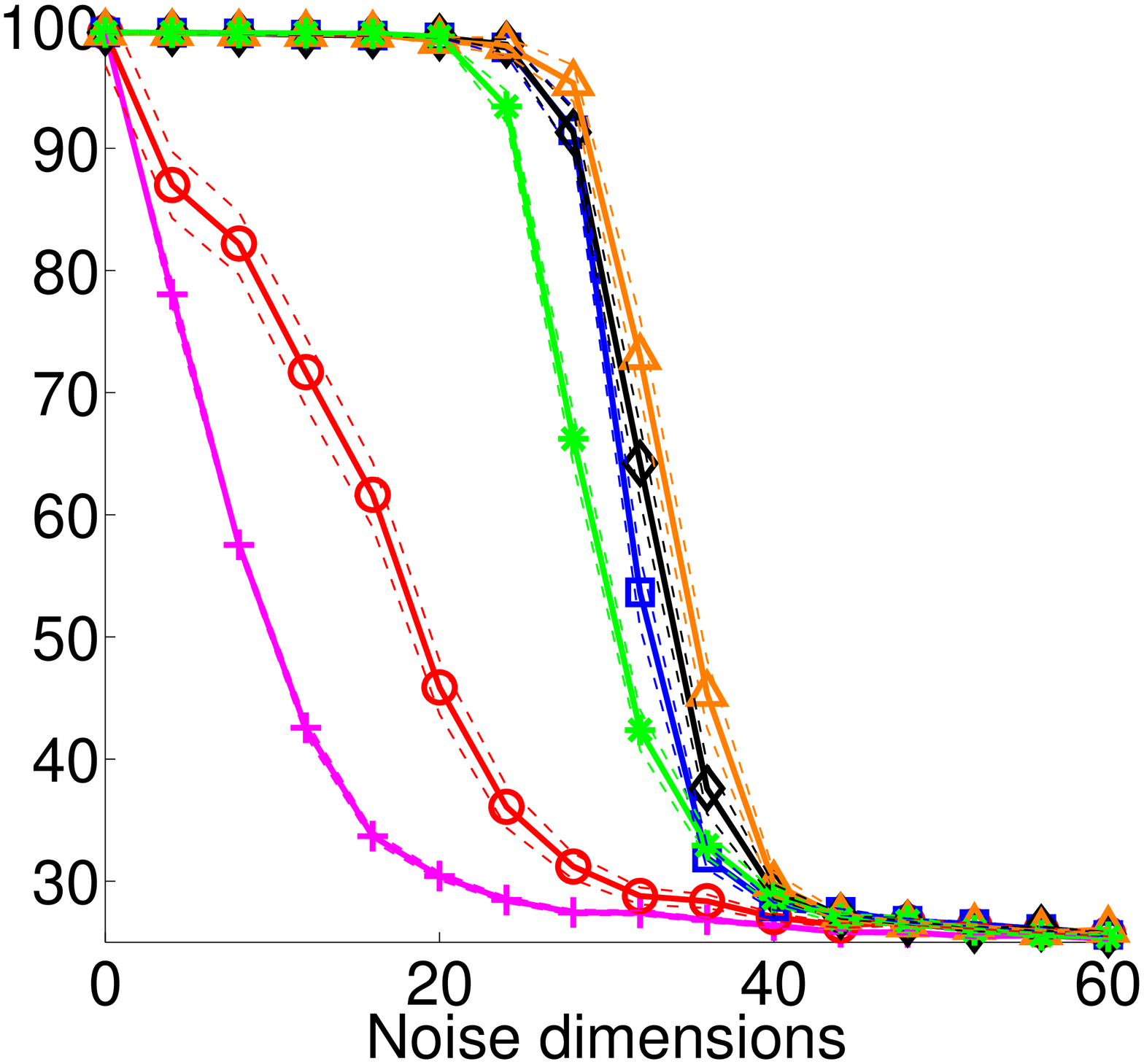}\\
		\includegraphics[width=12cm]{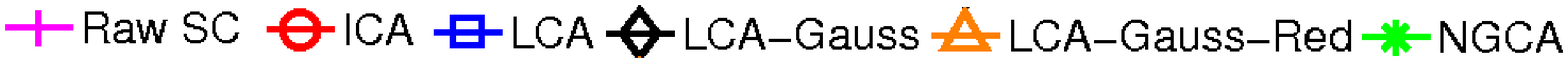}
	\caption{Average clustering accuracy (100\%  = perfect clustering,
chance is 50\% for the first two datasets and 20\% for the last one)
on 100 runs for varying number
of dimensions of noise added. The error bars represent one standard error.
Left: mixture of isotropic Gaussians presented in
Fig.~(\ref{fig:two_blob_data}) (left). Centre: two concentric
circles presented in Fig.~(\ref{fig:two_blob_data}) (centre). Right:
mixture of five Gaussians presented in Fig.~(\ref{fig:two_blob_data}) (right).
	\label{fig:cluster_error}}
\end{center}
\end{figure*}

We first try to solve three clustering problems: one for which the clusters
are convex and the direction of interest does not have a Gaussian marginal (Fig.~(\ref{fig:two_blob_data}), left), one
for which the clusters are not convex (Fig.~(\ref{fig:two_blob_data}), middle), and one for which the directions
of interest have almost Gaussian marginals
(Fig.~(\ref{fig:two_blob_data}), right).
Following~\cite{BachH07,Bach2003_learningSC}, the data is progressively corrupted by adding dimensions of white Gaussian noise, then whitened. We compare here the clustering accuracy, which is defined as $\frac{100}{n} \min_P \tr(EP)$ where $E$ is the confusion matrix and $P$ is the set of permutations over cluster labels, obtained with  the following five techniques:
\begin{enumerate}
	\item Spectral clustering (SC)~\cite{Ng01onspectral} on the whitened data (using the code of~\cite{Chen10});
	\item SC on the projection on the first two components found by FastICA using the best contrast function and the correct number of components;
	\item SC on the data transformed using the metric learnt with LCA;
	\item SC on the data transformed using the metric learnt with the product of LCA and a Gaussian;
	\item SC on the projection of the data found using NGCA~\cite{Blanchard06} with the correct number of components.
\end{enumerate}
Our choice of spectral clustering stems from its higher clustering performance compared to $K$-means. Results are reported in Fig.~(\ref{fig:cluster_error}). Because of the whitening, the Gaussian components in the first dataset are shrunk along
the direction containing information. As a result, even with little noise added, the information gets swamped and spectral clustering fails completely. On the other hand, LCA and its variants are much more robust to the presence of irrelevant dimensions. Though NGCA works very well on the first dataset, where there is only one relevant component, its performance drops quickly when there are two relevant components (note that, for all datasets, we provided the true number of relevant dimensions as input to NGCA). This is possibly due to the deflation procedure which is not adapted when no single component can be clearly identified in isolation. This is in contrast with LCA and its variants which circumvent this issue, thanks to their global optimisation procedure. Note also that \emph{LCA-Gauss} allows us to perform \emph{unsupervised} dimensionality reduction with the same performance as previously proposed supervised algorithms (e.g.,~\cite{Bach2003_learningSC}).

\begin{figure*}[ht!]
\begin{center}
	\includegraphics[width=4.2cm]{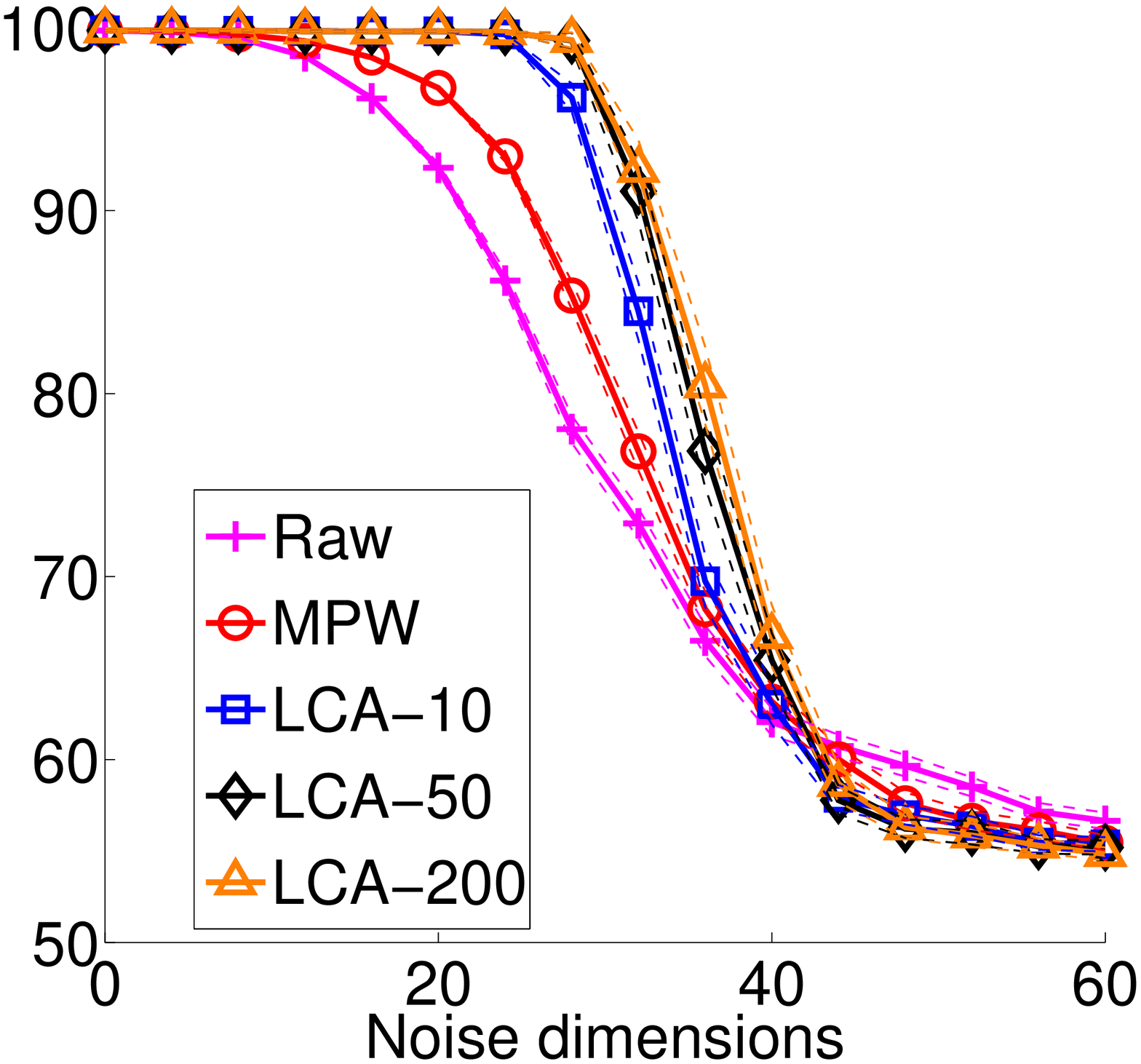}\hspace*{10mm}
	\includegraphics[width=4.2cm]{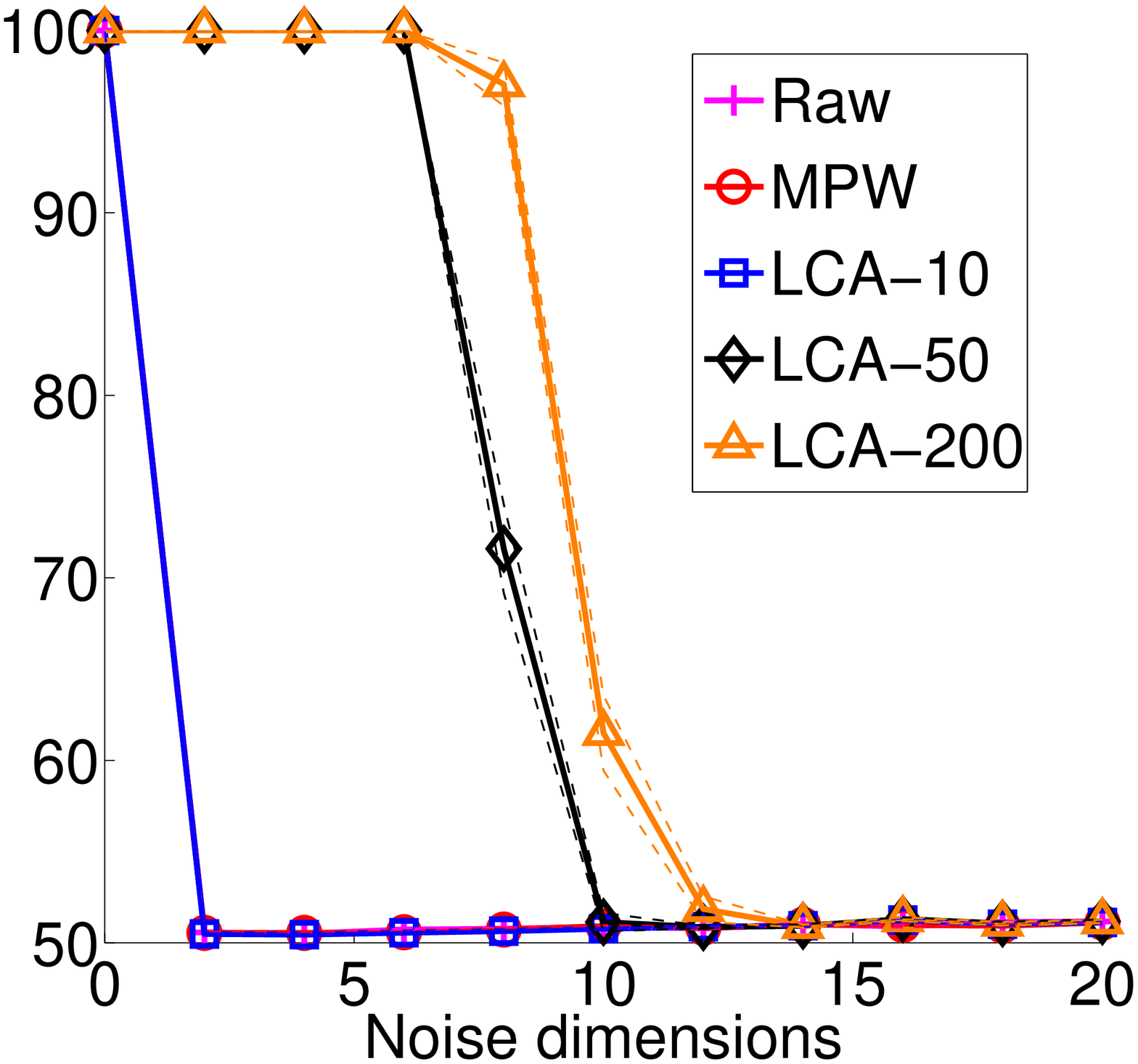}\hspace*{10mm}
	\includegraphics[width=4.7cm]{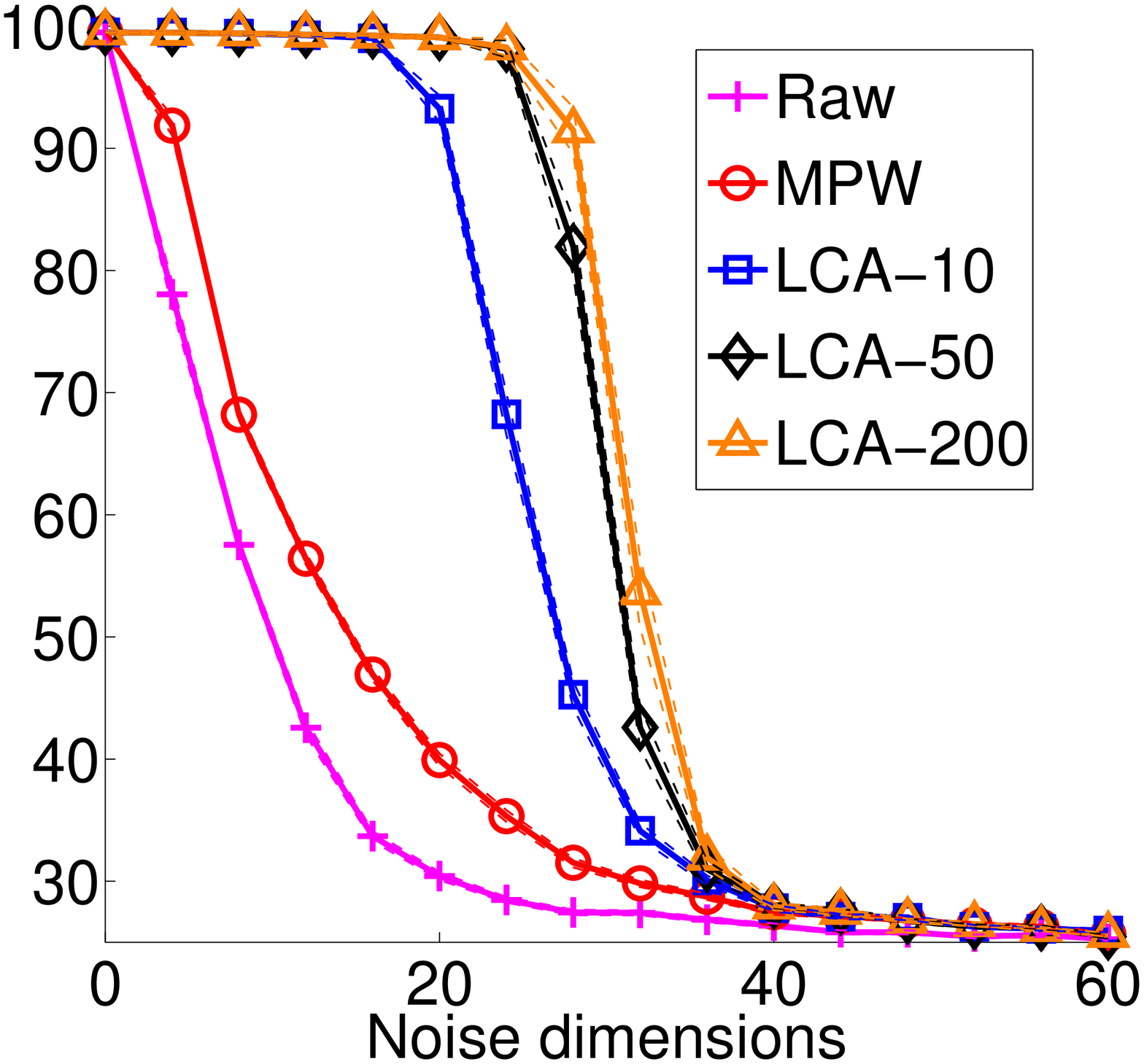}
	\caption{Average clustering accuracy (100\%  = perfect clustering,
chance is 50\% for the first two datasets and 20\% for the last one)
on 100 runs for varying number of dimensions of noise added and varying number of EM iterations in the LCA algorithm (MPW = one iteration). The error bars represent one standard error. Left: mixture of isotropic Gaussians presented in
Fig.~(\ref{fig:two_blob_data}) (left). Centre: two concentric
circles presented in Fig.~(\ref{fig:two_blob_data}) (centre). Right:
mixture of five Gaussians presented in Fig.~(\ref{fig:two_blob_data})
(right).
	\label{fig:comparison_mpq_lca}}
\end{center}
\end{figure*}

Figure~(\ref{fig:comparison_mpq_lca}) shows the clustering accuracy on the three datasets for various numbers of EM iterations, one iteration corresponding to Manifold Parzen Windows~\cite{Vincent02manifoldparzen} with a Gaussian kernel whose covariance matrix is the data covariance kernel. As one can see, running the EM algorithm to convergence yields a significant improvement in clustering accuracy. The performance of Manifold Parzen Windows could likely have been improved with a careful initialisation of the original kernel, but this would have been at the expense of the simplicity of the algorithm.

\subsection{LCA as a density model}
We now assess the quality of LCA as a density model. We build a density
model of the USPS digits dataset, a 256-dimensional dataset of handwritten digits.
We compared several algorithms:
\begin{itemize}
	\item An isotropic Parzen windows estimator with the bandwidth
estimated using LCA (replacing $\Sigma^*$ of Eq.~(\ref{eq:optimal_sigma}) by
$\lambda I$ so that the two matrices have the same trace);
	\item A Parzen windows estimator with diagonal metric (equal to
the diagonal of $\Sigma^*$ in Eq.~(\ref{eq:optimal_sigma});
	\item A Parzen windows estimator with the full metric as obtained
using LCA;
	\item A single Gaussian model;
	\item A product of a Gaussian and a Parzen windows estimator (as described in Section~\ref{sec:mult_gauss_components}).
\end{itemize}
The models were trained on a set of 2000 datapoints and regularized by
penalizing the trace of~$\Sigma^{-1}$ (in the case of the last model, both
covariance matrices, local and global, were penalized). The regularisation
parameter was optimized on a validation set of 1000 datapoints. For the last
model, the regularisation parameter of the global covariance was set to the
one yielding the best performance for the full Gaussian model on the
validation set. Thus, we only had to optimize the regularisation parameter
for the local covariance.

The final performance was then evaluated on a set of 3000 datapoints which
had not been used for training nor validation. We ran the experiment 20
times, randomly selecting the training, validation and test set each time.

Fig.~(\ref{fig:compare_parzen_usps}) shows the mean and the standard error
of the negative log-likelihood on the test set. As one can see, modelling
all dimensions using the Parzen windows estimator leads to poor performance
in high dimensions, despite the regulariser and the leave-one-out criterion.
On the other hand, LCA-Gauss and LCA-Gauss-Red clearly outperform all the
other models, justifying our choice of modelling some dimensions using a Gaussian. Also, as opposed to the previous experiments, there is no
performance gain induced by the use of LCA-Gauss-Red as opposed to
LCA-Gauss, which we believe stems from the fact that the switch from one
model to the other is easier to make when there are plenty of dimensions to
choose from. The poor performance of LCA-Full is a clear indication of the problems suffered by Parzen windows in high dimensions.

\begin{figure}[htbp]
\begin{center}
	%\subFig.{
	%\raisebox{-2.8cm}{
	%\includegraphics[width=5.5cm]{lca_usps_train.eps}}}
	%\subtable{
\begin{tabular}{|c|r|}
\hline
LCA - Isotropic 	&$ 269.78 \pm     0.18$ \\
\hline
LCA - Diagonal	& $   109.59 \pm     0.56$ \\
\hline
LCA - Full			&$  32.98 \pm     0.35$ \\
\hline
Gaussian 		&$  32.27 \pm     0.36$ \\
\hline
LCA - Gauss		&$ 19.09 \pm     0.39$ \\
\hline
LCA - Gauss - Red	&$  19.09\pm     0.39$ \\
\hline
\end{tabular}
%}
	\caption{Test negative log-likelihood on the USPS digits dataset, averaged over 20 runs.
	\label{fig:compare_parzen_usps}}
\end{center}
\end{figure}

\subsection{Subsampling}
\begin{figure*}[ht!]
\begin{center}
%\hspace*{-.3cm}
\begin{tabular}{|c||r|r|r||r|r|r|}
\hline
&\small{	N = 1000	}&\small{ N = 3000	}&\small{ N = 6000}&\small{	N = 1000	}&\small{ N = 3000	}&\small{ N = 6000}\\
\hline\hline
\small{B = 1000		}&\small{   $6.43 \pm     0.10$ }&\small{$   2.70 \pm     0.06$ }&\small{$    -0.10 \pm     0.03$ }&\small{$   6.43 \pm     0.10$ }&\small{$   2.80 \pm     0.06$ }&\small{$    -0.17 \pm     0.02$ }\\
\hline
\small{B = 3000		}&\small{   $6.58 \pm     0.07$ }&\small{$   2.73 \pm     0.05$ }&\small{$    -0.06 \pm     0.03$ }&\small{$   6.54 \pm     0.07$ }&\small{$   2.80 \pm     0.06$ }&\small{$    -0.11 \pm     0.02$ }\\
\hline
\small{B = 6000		}&\small{   $6.22 \pm     0.08$ }&\small{$   2.21 \pm     0.03$ }&\small{$   0.00 \pm     0.01$ }&\small{$   6.18 \pm     0.07$ }&\small{$   1.98 \pm     0.03$ }&\small{$   0.02 \pm     0.02$ }\\
\hline
\multicolumn{1}{r}{}\\
\hline
&\small{	N = 1000	}&\small{ N = 3000	}&\small{ N = 6000}&\small{	N = 1000	}&\small{ N = 3000	}&\small{ N = 6000}\\
\hline\hline
\small{B = 1000}&\small{$   2.12 \pm     0.16$ }&\small{$   0.20 \pm     0.08$ }&\small{$   0.65 \pm     0.05$}&\small{$   2.01 \pm     0.17$ }&\small{$   0.07 \pm     0.09$ }&\small{$   0.15 \pm     0.03$ }\\
\hline
\small{B = 3000}&\small{$   2.11 \pm     0.16$ }&\small{$   0.30 \pm     0.07$ }&\small{$   0.46 \pm     0.04$}&\small{$   2.01 \pm     0.17$ }&\small{$   0.16 \pm     0.09$ }&\small{$   0.09 \pm     0.03$ }\\
\hline
\small{B = 6000}&\small{$   1.54 \pm     0.16$ }&\small{$    -0.75 \pm     0.07$ }&\small{$    -0.01 \pm     0.01$}&\small{$   1.43 \pm     0.15$ }&\small{$    -1.07 \pm     0.08$ }&\small{$   0.01 \pm     0.02$ }\\
\hline
\end{tabular}
\caption{Train (top) and test (bottom) negative log-likelihood differences induced by the use of
smaller batch and neighbourhood sizes compared to the original model
($\gamma = 0$, $B = 6000$, $N = 6000$) for $\gamma = 0.3$ (left) and $\gamma
= 0.6$ (right). A negative value means better
performance.\label{fig:subsample_test}}
\end{center}
\end{figure*}
We now evaluate the loss in performance incurred by the use the subsampling procedure described in
Section~\ref{sec:subsampling}, both on the train and test negative log-likelihoods. For
that purpose, we used the USPS digit recognition dataset, which contains
8298 datapoints in dimension 256, which we randomly split into a training
set of $n=6000$ datapoints, using the rest as the test set. We
tested the following hyperparameters:
\bit
  \item Discount factor $\gamma = 0.3, 0.6, 0.9$ ,
  \item Batch size $B = 1000, 3000, 6000$ ,
  \item Neighbourhood size $N = 1000, 3000, 6000$ .
\eit

Fig.~(\ref{fig:subsample_test})
show the log-likelihood differences induced by the use of smaller batch
sizes and neighbourhood sizes. For each set of hyperparameters, 20
experiments were run using different training and test sets, and the means
and standard errors are reported. The results for $\gamma = 0.9$ were very
similar and are not included due to space constraints.

Three observations may be made. First, reducing the batchsize has little
effect, except when $\gamma$ is small. Second,  reducing the neighbourhood
size has a regularizing effect at first but drastically hurts the
performance if reduced too much. Third, the value of $\gamma$, the discount factor, has little
influence, but larger values proved to yield more consistent test
performance, at the expense of slower convergence. The consistency of these
results shows that it is safe to use subsampling (with values of $\gamma =
0.6$, $B = 100$ and $N = 3000$, for instance) especially if the training
set is very large.

\section{Conclusion}
Despite its importance, the learning of local or global metrics is usually an
overseen step in many practical algorithms. We have proposed an extension
of the general bandwidth selection problem to the multidimensional case,
with a generalisation to the case where several components are Gaussian.
Additionally, we proposed an approximate scheme suited to large datasets
which allows to find a local optimum in linear time. We believe LCA can be
an important preprocessing tool for algorithms relying on local distances,
such as manifold learning methods or many semi-supervised algorithms.
Another use would be to cast LCA within the mean-shift algorithm, which
finds the modes of the Parzen windows estimator, in the context of image
segmentation~\cite{Comaniciu02meanshift}. In the future, we would like to
extend this model to the case where the metric is allowed to vary with the
position in space, to account for more complex geometries in the dataset.

\subsection*{Acknowledgements}
Nicolas Le Roux and Francis Bach are supported in part by the European Research Council (SIERRA-ERC-239993). We would also like to thank Warith Harchaoui for its valuable input.

\newpage
\bibliography{local_component_analysis}
\bibliographystyle{plain}

\section*{Appendix}
We prove here Proposition~\ref{prop:mult_gauss}.
\begin{proof}
If $M_1$ is singular, then the minimum value is $-\infty$, because we can
have $B_G^\top M_1 B_G$ bounded while $B_G B_G^\top$ tends to $+ \infty$
(for example, if $d_1=1$, and $u_1$ is such that $M_1 u_1=0$, select $B_G =
\lambda u_1$ with $\lambda \to + \infty$). The reasoning is similar for
$M_2$.

We thus assume that $M_1$ and $M_2$ are invertible. We consider the eigendecomposition of $M_1^{-1/2} M_2 M_1^{-1/2} = U \Diag(e) U^\top$,
which corresponds to the generalized eigendecomposition of the pair
$(M_1,M_2)$.

Denoting $A_2 = U^\top M_1^{1/2} B_L$ and  $A_1 = U^\top M_1^{1/2} B_G$,
we have:
\beqas
 & &\tr B_G^\top M_1 B_G + \tr B_L^\top M_2 B_L
 - \log \det ( B_G B_G^\top + B_L B_L^\top) \\
  & = &  \tr A_1^\top A_1 + \tr A_2^\top   \Diag(e)  A_2 \\
 & &
 - \log \det ( A_1 A_1^\top + A_2 A_2^\top )   +  \log \det M_1 \\
 & = &  \tr A_1^\top A_1 + \tr A_2^\top   \Diag(e)  A_2 - \log \det (
A_2^\top A_2 )\\
&& - \log \det (A_1^\top ( \idm - A_2 (A_2^\top A_2)^{-1} A_2^\top ) A_1)
    +  \log \det M_1 \; .
 \eeqas

By taking derivatives with respect to $A_1$,
we get
\beq
A_1 = (\idm - \Pi_2) A_1 (A_1^\top ( \idm - \Pi_2 ) A_1)^{-1} \; ,
\label{eq:A_1}
\eeq
with $\Pi_2 = A_2 (A_2^\top A_2)^{-1} A_2^\top$. By left-multiplying both sides of Eq.~(\ref{eq:A_1}) by $A_2^\top$, we obtain
\[
A_2^\top A_1 = 0 \; .
\]
By left-multiplying by $A_1^\top$, we get
\[
A_1^\top A_1 = \idm \;.
\]
Thus, we now need to minimize with respect to $A_2$ the following
cost function

\beqas
d_1 + \tr A_2^\top   \Diag(e)  A_2
 - \log \det (A_2^\top A_2)
    +  \log \det M_1
 \eeqas

Let $s$ be the vector of singular values of $A_2$, ordered in \emph{decreasing}
order and let the $e_i$ be ordered in \emph{increasing} order.
We have:
\[
\tr \Diag(e) A_2 A_2^\top  = - \tr(-\Diag(e) A_2 A_2^\top)\geqslant \sum_{i} e_i s_i^2\;,
\]
 with
equality if and only if the eigenvectors of $A_2 A_2^\top$ are aligned with
the ones of $\Diag(e)$ (the $-e_i$ being also in decreasing order) (Theorem 1.2.1,~\cite{Borwein00convexanalysis}).

Thus, we have $A_2 A_2^\top = \diag(s)^2$ with only $d_2$ non-zero elements
in $s$. Let $J_2$ be the index of non zero-elements. We thus need to
minimize
$$
d_1 + \log \det M_1   + \sum_{j \in J_2} ( e_j s_j^2  - \log s_j^2 )\; ,
$$
with optimum $s_j^2 = e_j^{-1}$ and value:
$$
d_1 + d_2 + \log \det M_1 + \sum_{j \in J_2} \log e_j \; .
$$
Thus, we need to take $J_2$ corresponding to the smallest eigenvalues $e_j$. If
we also optimize with respect to $d_2$, then $J_2$ must only contain the elements smaller than 1.
\end{proof}

\end{document}